\documentclass[conference]{IEEEtran}

\usepackage[]{cite}
\usepackage{array}
\usepackage{amssymb,amsmath,caption}
\usepackage{graphicx}
\usepackage{psfrag}
\usepackage{algorithmic}
\usepackage{algorithm}
\usepackage{varioref}
\usepackage{booktabs}
\usepackage{tabularx}
\usepackage{caption,subcaption}
\usepackage{subfig}

\usepackage{float}

\usepackage{url}
\begin{document}
\onecolumn

\title{Feature Selection for classification of hyperspectral data by minimizing a tight bound on the VC dimension}

\author{
    \IEEEauthorblockN{Phool Preet\IEEEauthorrefmark{1}\IEEEauthorrefmark{3}, Sanjit Singh Batra\IEEEauthorrefmark{2}\IEEEauthorrefmark{3} and Jayadeva\IEEEauthorrefmark{1}\IEEEauthorrefmark{4}}\\
    \IEEEauthorblockA{\IEEEauthorrefmark{1}Department of Electrical Engineering, Indian Institute of Technology Delhi}
    \IEEEauthorblockA{\IEEEauthorrefmark{2}Department of Computer Science, Indian Institute of Technology Delhi\\ }
    \IEEEauthorblockA{\IEEEauthorrefmark{3}Equally contributing authors\\ }
    \IEEEauthorblockA{\IEEEauthorrefmark{4}Corresponding author: jayadeva@ee.iitd.ac.in \\ }

}


%


\maketitle

\begin{abstract}
Hyperspectral data consists of large number of features which require sophisticated analysis to be extracted. A popular approach to reduce computational cost,  facilitate information representation and accelerate knowledge discovery is to eliminate bands that do not improve the classification and analysis methods being applied. In particular, algorithms that perform band elimination should be designed to take advantage of the specifics of the classification method being used. This paper employs a recently proposed filter-feature-selection algorithm based on minimizing a tight bound on the VC dimension. We have successfully applied
this algorithm to determine a reasonable subset of bands without any user-defined stopping criteria on widely used hyperspectral images and demonstrate that this method outperforms state-of-the-art methods in terms of both sparsity of feature set as well as accuracy of classification.\end{abstract}


%
\IEEEpeerreviewmaketitle

\section{Introduction}
In recent years, hyperspectral image analysis has gained widespread importance among the remote sensing community. Hyperspectral sensors capture image data over hundreds of contiguous spectral channels (termed as bands), covering a broad spectrum of wavelengths(0.4-2.5 $\mu$m). Each hyperspectral image's scene is represented as an image cube. Hyperspectral data is increasingly becoming a valuable tool in many areas such as agriculture, mineralogy, surveillance, chemical imaging and automatic target recognition. A common task in such applications is to classify hyperspectral images. The abundance of information provided by hyperspectral data can be leveraged to enhance the ability to classify and recognize materials. However, the high dimensionality of hyperspectral data presents several obstacles. Firstly, it increases the computational complexity of classification. Further, it has been noted that highly correlated features have a negative impact on classification accuracy \cite{PAL}. Another quandary often observed in the classification of hyperspectral data is the \textit{Hughes effect}, which posits that in the presence of a limited number of training samples, the addition of features may have a considerable negative impact on the accuracy of classification. Therefore, dimensionality reduction is often employed in hyperspectral data analysis to reduce computational complexity and improve classification accuracy.

Dimensionality reduction methods can be divided into two broad categories: \textbf{feature extraction} and \textbf{feature selection}. Feature extraction methods, which transform the original data into a projected space, include for instance, projection pursuit(PP)\cite{Agustin}, principal component analysis(PCA)\cite{Abhishek_Agarwal} and independent component analysis(ICA)\cite{SEN_JIA}. Feature selection methods, on the other hand, attempt to identify a subset of features that contain the fundamental characteristics of the data. Most of the unsupervised feature selection methods are based on \textit{feature ranking}, which construct and evaluate an objective matrix based on various criteria such as information divergence \cite{Chein_Chang_ID}, maximum-variance principal component analysis (MVPCA)\cite{Chein_Chang}, and mutual information (MI) \cite{Baofeng_Guo}.

This paper explores the application of a novel feature selection method based on minimizing a tight bound on the VC dimension\cite{FS_Jayadeva}, on hyperspectral data analysis. We present a comparison with various state-of-the-art feature selection methods on benchmark hyperspectral datasets. We used the Support Vector Machine (SVM) classifier\cite{Farid_Melgani} to assess the classification accuracy, following feature selection. Rest of the paper is organized as follows. Section II briefly describes the related work and background. In section III, the various feature selection methods used in this paper are described. Section IV describes the datasets used, the experimental setup and the results obtained on benchmark hyperspectral datasets.

\section{Background and Related Work}

Dimensionality reduction prior to classification is advantageous in hyperspectral data analysis because the dimensionality of the input space greatly affects the performance of many supervised classification methods\cite{Baofeng_Guo}. Further, there is a high likelihood of redundancy in the features and it is possible that some features contain less discriminatory information than others. Moreover, the high-dimensionality imposes requirements for storage space and computational load. The analysis in\cite{PAL} supports this line of reasoning and suggests that feature selection may be a valuable procedure in preprocessing hyperspectral data for classification by the widely used SVM classifier.

In hyperspectral image analysis, feature selection is preferred over feature extraction for dimensionality reduction\cite{PAL},\cite{Adolfo_Martinez}. Feature extraction methods involve transforming the data and hence, crucial and critical information may be compromised and distorted. In contrast, feature selection methods strive to discover a subset of features which capture the fundamental characteristics of the data, while possessing sufficient capacity to discriminate between classes. Hence, they have the advantage of preserving the relevant original information of the data.

There are various studies which establish the usefulness of feature selection in hyperspectral data classification. \cite{PAL} lists various feature selection methods for hyperspectral data such as the SVM Recursive Feature Elimination (SVM-RFE) \cite{Guyon}, Correlation based Feature Selection(CFS) \cite{Hall}, Minimum Redundancy Maximum Relevance(MRMR) \cite{Peng} feature selection and Random Forests \cite{Breiman}. In \cite{Chein_Chang}, a band prioritization scheme based on Principal Component Analysis (PCA) and classification criterion is presented. Mutual information is a widely used quantity in various feature selection methods. In a general setting, features are ranked based on the mutual information between the spectral bands and the reference map(also known as the \textit{ground truth}). In \cite{Baofeng_Guo}, mutual information is computed using the estimated reference map obtained by using available \emph{a priori} knowledge about the spectral signature of frequently-encountered materials.

Recently, Brown et al\cite{Gavin_Brown} have presented a framework for unifying many information based feature selection selection methods. Based on their results and suggestions we have chosen the set of feature selection methods that they recommend outperform others, in various situations, which are elaborated in the next section for the purposes of our analysis. In \cite{FS_Jayadeva} a feature selection method based on minimization of a tight bound on the VC dimension is presented. This paper presents the first application of this novel method to hyperspectral data analysis.

\section{Feature Selection Methods}

\subsection{PCA based Feature Selection}

\noindent Principal Component Analysis (PCA) is one of the most extensively used feature selection method. It transforms the data in such a way that the projections of the transformed data(termed as the \textit{principal components}) exhibit maximal variance. Chein et al.\cite{Chein_Chang} presents a band prioritization method based on Principal Component Analysis. For our experiments, we consider the features obtained from PCA to be the eigenvectors sorted by their corresponding eigenvalues.

\subsection{Information Theoretic Feature Selection}

\noindent Feature selection techniques can be broadly divided into two categories: \textbf{classifier-dependent}(€˜wrapper€™ and €˜embedded€™ methods) and \textbf{classifier-independent} (€˜filter€™ methods). \textit{Wrapper methods} rank the features based on the accuracy of a particular classifier. They have the disadvantage of being computationally expensive and classifier-specific. \textit{Embedded methods} exploit the structure of particular classes of learning models to guide the feature selection process. In contrast, \textit{Filter methods} evaluate statistics of the data, independent of any classifier and define a heuristic scoring criterion(\textit{relevance index}). This scoring criterion is a measure of how useful a feature can be, when input to a classifier. 

\subsubsection*{\hspace{-14 pt}MRMR}
The \emph{Minimum-Redundancy Maximum-Relevance} criterion was proposed by Peng et al. \cite{Peng}. Let $X$ be our feature vector and $Y$ is training label then mRMR criterion is given by \\
\begin{equation}
J_{mrmr}(X_k) = I(X_k;Y) - \frac{1}{|S|}\sum_{j\in S} I(X_k;X_j) 
\end{equation}
$I(\cdot;\cdot)$ denotes the mutual information and $S$ is subset of selected features. Feature $X_k$ is ranked on the basis of mutual information between $X_k$ and training labels in order to maximize the \textit{relevance} and also on the basis of the mutual information between $X_k$ and already selected features $X_j$ (where $j \in S$) in order to minimize the \textit{redundancy}.

\subsubsection*{\hspace{-14 pt}JMI}
Yang et al. in \cite{Yang} proposed \emph{Joint Mutual Information}.\\
\begin{equation}
J_{jmi}(X_k) = \sum_{j\in S} I(X_kX_j;Y) 
\end{equation}
It is defined as the mutual information between the training labels and a joint random variable $X_kX_j$. It ranks the features $X_k$ on the basis of how complimentary it is with already selected features $X_j$. 

\subsubsection*{\hspace{-14 pt}CMIM}
\emph{Conditional Mutual Information Maximization} is another information theoretic criterion that was proposed by Fleuret \cite{Fleuret}.
\begin{equation}
J_{cmim}(X_k) = arg max_k \{\min_{j\in S} [I(X_k;Y|X_j)] \}
\label{eq_cmim}
\end{equation}
\noindent The feature which maximizes the criterion in equation \ref{eq_cmim} at each stafe is selected as the next candidate feature. As a result, this criterion selects the feature that carries most information about $Y$ and also considers whether this information has been captured by any of the already selected features.

\subsection{RELIEF}
\noindent Relief is a feature weight based algorithm statistical feature selection method proposed by Kira and Rendell \cite{Kira}. Relief detects those features which are statistically relevant to the target concept. The algorithm starts with a weight vector $W$ initialized by zeros. At each iteration, the algorithm takes the feature vector $X_k$ belonging to a random instance and the feature vectors of the instance closest to $X_k$, from each class. The closest same-class instance is termed as a \textit{near-hit} and the closest different-class
instance is called a \textit{near-miss}. The weight vector is then updated using equation \ref{relief}. \\
\begin{equation}
W_i = W_i - (x_i-nearHit)^2 + (x_i-nearMiss)^2
\label{relief}
\end{equation} \\
Thus the weight of any given feature decreases if it differs from that feature's value in nearby instances of the same class \textbf{more} than nearby instances of the other class, and increases in the converse case. Features are selected if their relevance is greater than a defined threshold. Features are then ranked on the basis of their \textit{relevance}.

\subsection{Feature Selection by VC Dimension Minimization}
In order to perform feature selection via MCM, we solve the following linear programming problem:\\

\begin{gather}
\operatorname*{Min}_{w, b, h} ~~h + C \cdot \sum_{i = 1}^M q_i \label{obj5}\\
h \geq y_i \cdot [{w^T x^i + b}] + q_i, ~i = 1, 2, ..., M \label{cons51}\\
y_i \cdot [{w^T x^i + b}] + q_i \geq 1, ~i = 1, 2, ..., M \label{cons52} \\
q_i \geq 0, ~i = 1, 2, ..., M. \label{cons53}
\end{gather}\\

where $x^i$, $~i = 1, 2, ..., M$ are the input data points and $y_i, ~i = 1, 2, ..., M$ are the corresponding target labels.

The classifier generated by the solving the above problem minimizes a tight bound on the VC dimension and hence yields a classifier that uses a small number of features\cite{FS_Jayadeva}\cite{Classifier_MCM}\cite{Classifier_MCM_arxiv}\cite{Regressor_MCM}\cite{Regressor_MCM_arxiv}. Here, the choice of $C$ allows a tradeoff between the complexity (machine capacity) of the classifier and the classification error.\\

Once $w$ and $b$ have been determined, to obtain a feature ranking, features are sorted in descending order based on the value of $\left|w^j\right|$ for each feature $j=1,2...D$.

\section{Experimental Setup and Results}

\noindent To assess the classification accuracy for the multi-class datasets in this paper, we use the "one-vs-rest" strategy. Each class is classified using the data belonging to rest of the classes as negative training samples. A Support Vector Machine classifier \cite{LibSVM} with an RBF kernel is used for classification. The box-constraint parameter of SVM, C, is set to a high value to give more emphasis on correct classification; the width of the Gaussian kernel is selected empirically.

\noindent To assess the ability of the different methods to pick out the best features in the scarcity of training data, we evaluate classification results for a fixed test/train ratio while varying the number of features output by the different methods. Number of bands selected using different methods are 1, 2, 3, 4, 5, 6, 7, 8, 9, 10, 13, 15, 20, 25, 30, 35, 40, 45 and 50.

\noindent Further, to asses the impact of the availability labeled data for training the model, we also evaluate the results for varying test/train ratios, while fixing the number of features. Different test/train ratio chosen for the experiment are
0.7, 0.75, 0.8, 0.85, 0.90 and 0.95.

\noindent In the one-vs-rest strategy, data often become highly unbalanced and hence accuracy(percentage of correctly classified points) alone is not a good metric of classification performance. Hence, we measure the Matthews Correlation Coefficient (MCC) for each class and computed the weighted MCC, where the weight of a class is derived from the fraction of training samples present in the one-vs-rest class split for that particular class. Matthews Correlation Coefficient is generally regarded as a balanced measure which can be used even if the classes are of very different sizes. Matthews Correlation Coefficient is given by equation~(\ref{eq_mcc})
\begin{equation}
mcc = \frac{tp * tn - fp * fn}{( tp + fp ) * ( tp + fn ) * ( tn + fp ) * ( tn + fn )} 
\label{eq_mcc}
\end{equation}

\noindent where \emph{tp} (true positive) is the number of correctly classified positive samples, \emph{fp} (false positive) is the
number of negative samples classified as positive samples. \emph{tn} (true negative) is the correctly classified negative 
samples and \emph{fn} (false negative) is the number of positive samples classified as negative samples. 
MCC is computed for each class and weighted average is calculated using the number of samples in each class.

\subsection{ Indian Pines Data-set}
\begin{figure}[H]
  \begin{minipage}{.5\linewidth}
    \centering
    \includegraphics[scale=.7]{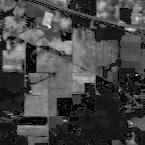}
    \subcaption{image of band 20 of Indian Pine}\label{figure_Indian_Pines_a}
  \end{minipage}%
  \begin{minipage}{.5\linewidth}
    \centering
    \includegraphics[scale=.7]{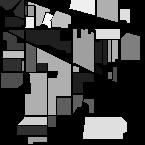}
    \subcaption{ground truth for Indian Pine}\label{figure_Indian_Pines_b}
  \end{minipage}
  \caption{\small{Indian Pines}}\label{figure_Indian_Pines}
\end{figure}

This scene was acquired by the AVIRIS sensor. Indian Pines is a 145$\times$145 scene containing 224 spectral reflectance bands in the wavelength range $0.4$--$2.5$ $10^{-6}$ meters. The Indian Pines scene contains two-thirds agriculture, and one-third forest or other natural perennial vegetation. A random band along with ground truth is shown in Figure \ref{figure_Indian_Pines}. The ground truth available is designated into sixteen classes. The corrected Indian Pines data-set contains 200 bands, obtained after removing bands covering the region of water absorption: (104-108), (150-163), 220.

\noindent Table \ref{table_Indian_Pines} gives the details of classes, number of samples and number of training and testing points for Indian Pines data-set.

\begin{table}[H]
  \centering
  \scriptsize
\begin{tabular}{ p{.04\columnwidth}  p{.35\columnwidth}  p{.1\columnwidth} p{.1\columnwidth} p{.1\columnwidth}  }
    \hline
    \#      &  Class 	                    &  Samples  & Train    & Test\\    \hline  
    1       &  Alfalfa                      &   46      &   3      &   43\\   
    2 	    &  Corn-notill 	                &   1428    &   134    &   1294\\   
    3 	    &  Corn-mintill 	            &   830     &   87     &   743\\   
    4  	    &  Corn 	                    &   237     &   28     &   209\\   
    5 	    &  Grass-pasture 	            &   483     &   42     &   441\\   
    6 	    &  Grass-trees 	                &   730     &   75     &   655\\   
    7 	    &  Grass-pasture-mowed          &	28      &	4      &   24\\   
    8 	    &  Hay-windrowed 	            &   478     &   54     &   424\\   
    9 	    &  Oats 	                    &   20      &   4      &   19\\   
    10 	    &  Soybean-notill 	            &   972     &   99     &   873\\   
    11 		&  Soybean-mintill 	            &   2455    &   285    &   2170\\   
    12 		&  Soybean-clean 	            &   593     &   68     &   525\\   
    13 		&  Wheat 	                    &   205     &   22     &   183\\   
    14 		&  Woods 	                    &   1265    &   120    &   1145 \\   
    15 		&  Buildings-Grass-Trees-Drives &	386     &	31     &   355\\   
    16 		&  Stone-Steel-Towers 	        &   93      &   9      &   84\\   \hline
  \end{tabular}
   \caption{\small{Different classes, number of samples and number of training and testing points
   in Indian Pines data-set}}\label{table_Indian_Pines}
\end{table}

\begin{figure}[H]
\captionsetup[subfigure]{labelformat=empty}
\begin{subfigure}{.5\textwidth}
\centering
\includegraphics[scale=.5]{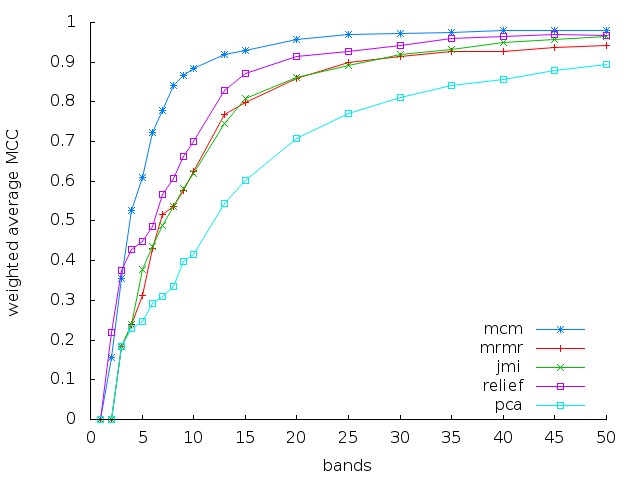}
\caption{\small Weighted average MCC vs number of bands plot for Indian Pines\\ dataset (test/train ratio = 0.90)}
\end{subfigure}
\begin{subfigure}{.5\textwidth}
\centering
\includegraphics[scale=.5]{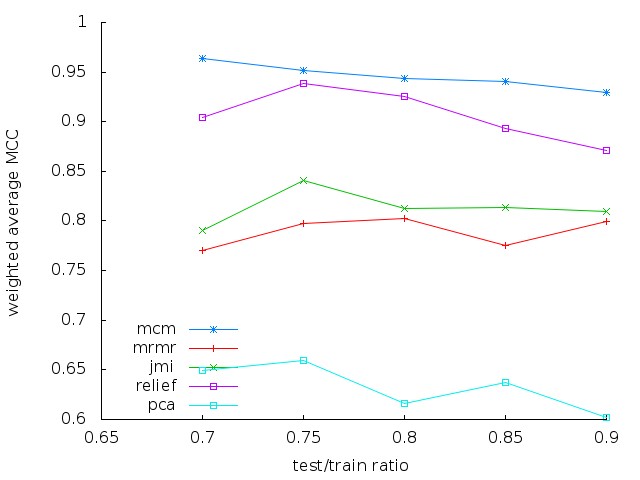}
\caption{\small Weighted average MCC vs test/train ratio plot for Indian Pines\\ dataset (bands = 15)}
\end{subfigure}
\caption{Indian Pines dataset}
\label{plot_Indian_Pines}
\end{figure}

 Figure \ref{plot_Indian_Pines} shows the plot of number of bands vs classification accuracy 
 (Matthews Correlation Coefficient) for Indian Pines data-set. This plot was generated using test/train ratio of 0.90 (Table \ref{table_Indian_Pines}). Plot of test-train ratio vs classification accuracy generated using first 15 bands is also shown in this figure.

\noindent 
Table \ref{table2_Indian_Pines} reports the indices of the top 15 selected bands by different feature selection methods.

\begin{table}[H]
  \centering
  \scriptsize
   \noindent \begin{tabular}{ p{.18\columnwidth}  p{.74\columnwidth} }
    \hline
    Band selection method  &  Fifteen selected bands \\  \hline
    MCM    &    114 133 118 129 70 127 125 131 46 116 128 63 134 159 53  \\
    MRMR   &    51 152 39 46 145 40 103 63 61 56 144 110 146 1 41   \\
    JMI    &	51 152 63 39 46 145 110 40 56 41 38 34 36 37 35   \\
    RELIEF &	96 86 94 97 30 85 84 61 83 90 32 89 200 78 108  \\
    PCA    &    180 183 182 196 179 172 194 177 178 185 187 176 174 175 181 \\ \hline
    \end{tabular}
    \caption{\small{comparison between selected bands using different band selection methods for Indian Pines data-set (test/train ratio-0.90) }}\label{table2_Indian_Pines}
\end{table}

Table \ref{table3_Indian_Pines} depicts the class-wise and weighted average accuracy measure.

\begin{table}[H]
  \scriptsize
  \centering
   \noindent \hspace{-10 pt} \begin{tabular}{ p{.29\columnwidth}  p{.1\columnwidth}  p{.1\columnwidth}  p{.1\columnwidth} p{.1\columnwidth} p{.1\columnwidth} }
    \hline 
    Type 	                  &  MCM        & MRMR 		& JMI       & RELIEF    & PCA  \\    \hline 
    Alfalfa                   &  0.820602   & 0.171495  & 0.652584  & 0.287088  & 0.436014  \\  
    Corn-notill 	          &  0.973721   & 0.839001  & 0.722158  & 0.789404  & 0.495645  \\ 
    Corn-mintill 	          &  0.934575   & 0.686495  & 0.645782  & 0.824429  & 0.360526  \\   
    Corn 	                  &  0.0667986  & 0.425368  & 0.278761  & 0.531958  & 0.0       \\   
    Grass-pasture 	          &  0.967502   & 0.894631  & 0.964059  & 0.859303  & 0.702175  \\            
    Grass-trees 	          &  0.923784   & 0.729554  & 0.761606  & 0.9737    & 0.405883  \\   
    Grass-pasture-mowed       &	 0.842576   & 0.674168  & 0.143084  & 0.539564  & 0.455965  \\   
    Hay-windrowed 	          &  0.933875   & 0.947756  & 0.954423  & 0.782468  & 0.626384  \\   
    Oats 	                  &  0.486037   & 0.30493   & 0.289147  & 0.519747  & 0.263709  \\
    Soybean-notill 	          &  0.86057    & 0.856167  & 0.849795  & 0.92911   & 0.457119  \\ 
    Soybean-mintill 	      &  0.991267   & 0.962822  & 0.96087   & 0.981014  & 0.894715  \\
    Soybean-clean 	          &  0.953628   & 0.497804  & 0.623142  & 0.628778  & 0.192737  \\  
    Wheat 	                  &  0.810033   & 0.696977  & 0.627184  & 0.941961  & 0.504204  \\  
    Woods 	                  &  0.990536   & 0.800397  & 0.891902  & 0.989561  & 0.79067   \\  
    Bldgs-Grass-Trees-Drives  &	 0.893755   & 0.531609  & 0.718466  & 0.735206  & 0.758364  \\             
    Stone-Steel-Towers 	      &  0.938078   & 0.289439  & 0.615549  & 0.485004  & 0.473028  \\ \hline
    Weighted Average		  &  0.929820 & 0.79932096 & 0.8093053 & 0.8714561 & 0.6023033 \\  \hline
  \end{tabular}
   \caption{\small{ MCC for different band selection methods for Indian Pines data-set (corresponding to  test/train ratio - 0.90, bands 15)}}\label{table3_Indian_Pines}
\end{table}

\subsection{Salinas} 
\begin{figure}[H]
  \begin{minipage}{.45\linewidth}
    \centering
    \includegraphics[width=80 pt, height=150 pt]{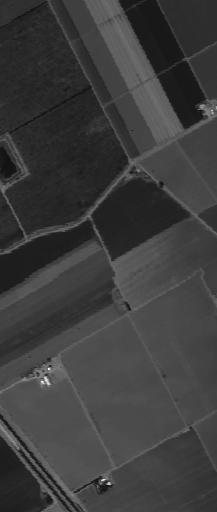}
    \subcaption{image for band 25 of Salinas}\label{figure_Salinas_a}
  \end{minipage}%
  \begin{minipage}{.45\linewidth}
    \centering
    \includegraphics[width=80 pt, height=150 pt]{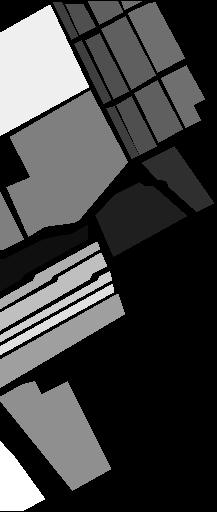}
    \subcaption{ground truth for Salinas}\label{figure_Salinas_b}
  \end{minipage}
  \caption{Salinas}\label{figure_Salinas}
\end{figure}

This scene was also gathered by the AVIRIS sensor and contains 224 bands. In this dataset, 20 water absorption bands [108-112], [154-167] and 224 have been removed during preprocessing. A random band along with the ground truth is shown in Figure \ref{figure_Salinas}. It includes vegetables, bare soils, and vineyard fields. Salinas' groundtruth consists of 16 classes. The dataset is available online \cite{Indian_Pines}.

\noindent Table \ref{table_Salinas} lists the different classes, number of samples and number of training and testing points in the Salinas dataset corresponding to test/train ratio of 0.90. 

\begin{table}[H]
  \scriptsize
  \centering
  \noindent \begin{tabular}{ p{.04\columnwidth}  p{.35\columnwidth}  p{.1\columnwidth} p{.1\columnwidth} p{.1\columnwidth} }
    \hline
    \#     &  Class 	                    &  Samples & Train 		& Test \\    \hline  
 	1 	    &  Brocoli green weeds 1        &  2009    &  194  		& 1815\\      
 	2 	    &  Brocoli green weeds 2 	    &  3726    &  357     	& 3369\\    
 	3 	    &  Fallow 	                    &  1976    &  193     	& 1783 \\    
 	4 	    &  Fallow rough plow 	        &  1394    &  142     	& 1252\\    
	5 	    &  Fallow smooth 	            &  2678    &  272     	& 2406\\    
 	6 	    &  Stubble 	                    &  3959    &  417     	& 3542\\    
 	7 	    &  Celery 	                    &  3579    &  367     	& 3212\\    
 	8 	    &  Grapes untrained 	        &  11271   &  1147    	& 10124\\    
 	9 	    &  Soil vinyard develop 	    &  6203    &  605     	& 5598\\    
 	10 		&  Corn senesced green weeds 	&  3278    &  365     	& 2913\\    
 	11 		&  Lettuce romaine 4wk 	        &  1068    &  119     	& 949\\    
 	12 		&  Lettuce romaine 5wk 	        &  1927    &  165     	& 1762\\    
 	13 		&  Lettuce romaine 6wk 	        &  916     &  95      	& 821\\    
 	14 		&  Lettuce romaine 7wk 	        &  1070    &  106     	& 964\\    
 	15 		&  Vinyard untrained 	        &  7268    &  735     	& 6533\\    
 	16 		&  Vinyard vertical trellis 	&  1807    &  175 		& 1632\\    \hline  
  \end{tabular}
   \caption{\small{Classes and number of test and training samples for Salinas (corresponding to  test/train ratio - 0.90)}}\label{table_Salinas}
\end{table}

\begin{figure}[H]
\captionsetup[subfigure]{labelformat=empty}
\begin{subfigure}{.5\textwidth}
\centering
\includegraphics[scale=.5]{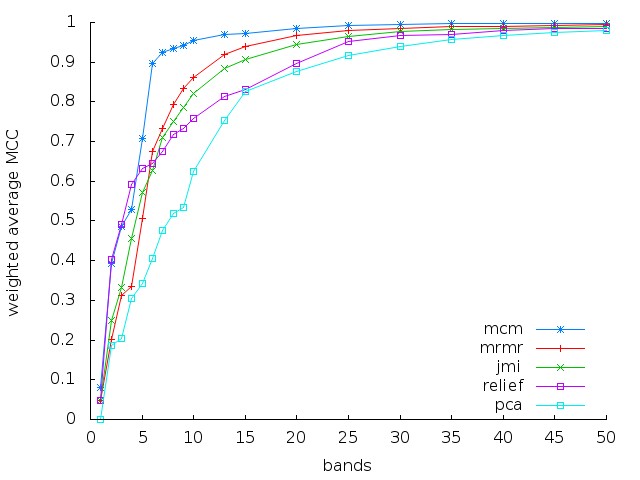}
\caption{\small Weighted average MCC vs number of bands plot for Salinas\\ dataset (test/train ratio - 0.90)}
\end{subfigure}
\begin{subfigure}{.5\textwidth}
\centering
\includegraphics[scale=.5]{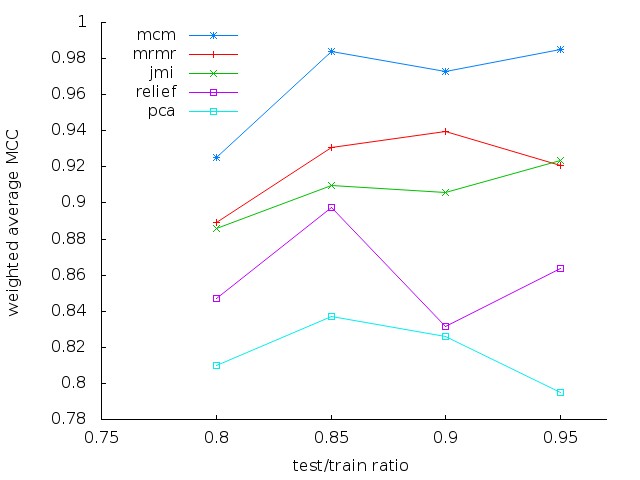}
\caption{\small Weighted average MCC vs test/train ratio plot for Salinas\\ dataset (bands - 15)}
\end{subfigure}
\caption{Salinas dataset}
\label{plot_Salinas}
\end{figure}

Number of bands vs classification accuracy plot is given in figure \ref{plot_Salinas}.
Plot of test-train ratio vs classification accuracy generated using first 15 bands is also shown in this figure.

Table \ref{table2_Salinas} shows the top 15 selected bands by different features selection methods.

\begin{table}[H]
  \centering
  \scriptsize
  \noindent \begin{tabular}{ p{.18\columnwidth}  p{.74\columnwidth} }
    \hline
    Band selection method  &  Fifteen selected bands \\  \hline
    MCM    &    76 68 66 57 73 90 133 41 43 56 74 60 89 75 123 \\
    MRMR   &    106 203 84 107 147 204 110 1 2 37 3 13 112 150 129   \\
    JMI    &	106 203 84 37 129 138 107 139 33 136 177 137 127 134 135   \\
    RELIEF &	43 46 45 44 201 47 48 52 51 50 42 198 49 53 54  \\
    PCA    &    188 195 185 192 184 201 189 187 190 199 179 194 183 202 193 \\ \hline
    \end{tabular}
    \caption{\small{comparison between selected bands using different band selection methods
    (test/train ratio-0.90)}}\label{table2_Salinas}
\end{table}

Table \ref{table3_Salinas} depicts the class-wise and weighted average accuracy measure.

\begin{table}[H]
  \scriptsize
  \centering
   \noindent \hspace{-15 pt} \begin{tabular}{ p{.32\columnwidth}  p{.1\columnwidth}  p{.1\columnwidth}  p{.1\columnwidth} p{.1\columnwidth} p{.1\columnwidth} }
    \hline 
    Type 	                     &  MCM           & MRMR 		& JMI        & RELIEF      & PCA  \\    \hline
   Brocoli green weeds 1        &  0.983902		& 0.877167      & 0.864284   &  0.934087   & 0.903624  \\
 	Brocoli green weeds 2 	    &  0.996648		& 0.919044      & 0.882694   &  0.982051   & 0.897363 \\       	Fallow 	                    &  0.961271		& 0.947172      & 0.654416   &  0.844062   & 0.542568  \\      	Fallow rough plow 	        &  0.96507		& 0.802292      & 0.804176   &  0.918218   & 0.496249  \\   
	Fallow smooth 	            &  0.998688		& 0.999563      & 0.999563   &  0.998689   & 0.999781  \\     
 	Stubble 	                &  0.979638		& 0.896754      & 0.917716   &  0.417727   & 0.989659  \\     
 	Celery 	                    &  0.988002		& 0.987175      & 0.810599   &  0.969313   & 0.890965  \\     
 	Grapes untrained 	        &  0.973939		& 0.969944      & 0.972389   &  0.974863   & 0.867423  \\     
 	Soil vinyard develop 	    &  0.955469		& 0.938005      & 0.969929   &  0.95706    & 0.891139  \\     
 	Corn senesced green weeds 	&  0.953872		& 0.973493      & 0.921169   &  0.880737   & 0.760172  \\     
 	Lettuce romaine 4wk 	    &  0.980071		& 0.885836      & 0.707766   &  0.838098   & 0.863744  \\     
 	Lettuce romaine 5wk 	    &  0.98101 		& 0.837188      & 0.791131   &  0.7315     & 0.679636  \\     
 	Lettuce romaine 6wk 	    &  0.984424		& 0.970165      & 0.928471   &  0.827189   & 0.749821  \\     
 	Lettuce romaine 7wk 	    &  0.972923		& 0.831091      & 0.723109   &  0.737297   & 0.603268  \\     
 	Vinyard untrained 	        &  0.960854		& 0.959079      & 0.958558   &  0.943936   & 0.914944  \\     
 	Vinyard vertical trellis 	&  0.959804		& 0.931492		& 0.838574	 &  0.952117   & 0			\\ 
 	\hline
 	Weighted Average			&  0.9727331    & 0.9396902     & 0.9057743  & 0.83178004  & 0.8259082 \\ \hline
  \end{tabular}
   \caption{\small{ MCC for different band selection methods for Salinas data-set (corresponding to  test/train ratio - 0.80, bands 15)}}\label{table3_Salinas}
\end{table}

\subsection{Botswana Data-Set}
\begin{figure}[H]
  \begin{minipage}{.45\linewidth}
    \centering
    \includegraphics[scale=.2]{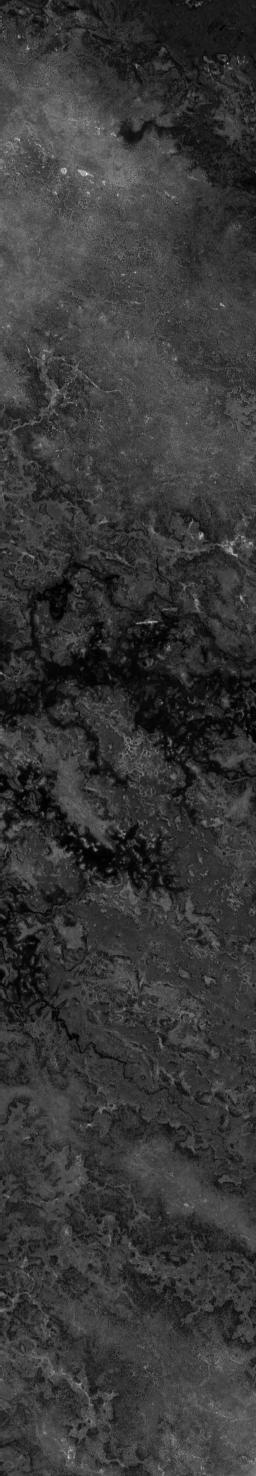}
    \subcaption{image for band 20 of Botswana}\label{figure_Botswana_a}
  \end{minipage}%
  \begin{minipage}{.45\linewidth}
    \centering
    \includegraphics[scale=.2]{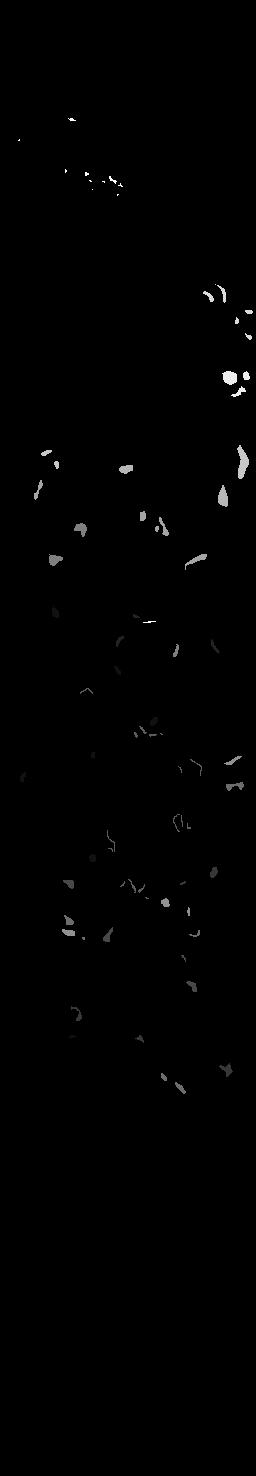}
    \subcaption{ground truth for Botswana}\label{figure_Botswana_b}
  \end{minipage}
  \caption{Botswana}\label{figure_Botswana}
\end{figure}

The Botswana dataset was acquired by the Hyperion sensor at 30m pixel resolution over a 7.7 km strip in 242 bands covering the 400-2500 nm portion of the spectrum in 10nm windows. Uncalibrated and noisy bands that cover water absorption features were removed, and the remaining 145 bands were included as candidate features\cite{Indian_Pines}. This dataset consists of observations from 14 identified classes representing the land cover types in seasonal swamps, occasional swamps, and drier woodlands. A random band along with ground truth for Botswana data-set is shown in Figure \ref{figure_Botswana}.

\noindent Table \ref{table_Botswana} gives the listing of number of samples and number of training and testing points in Botswana data-set corresponding to test train ratio 0.90. 

\begin{table}[H]
  \scriptsize
  \centering
   \noindent \begin{tabular}{ p{.08\columnwidth}  p{.3\columnwidth}  p{.12\columnwidth} p{.12\columnwidth} p{.12\columnwidth} }
    \hline 
    class   &  Type 	 			& Samples     &  Train    	& Test		\\    \hline 
    1 	    &  Water	 			& 270			&  35    	& 235    	\\      
 	2 	    &  Hippo grass 	 		& 101   		&  10     	& 91    	\\    
 	3 	    &  Floodplain grasses1	& 251          &  28     	& 223    	\\    
 	4 	    &  Floodplain grasses2 	& 215			&  20     	& 195     	\\    
	5 	    &  Reeds1 	 			& 269		&  29     	& 240    	\\    
 	6 	    &  Riparian 	 		& 269          &  34     	& 235    	\\    
 	7 	    &  Firescar2	 		& 259          &  24    	& 235    	\\    
 	8 	    &  Island interior 	 	& 203       	&  18    	& 185   	\\    
 	9 	    &  Acacia woodlands 	& 314   		&  31     	& 283    	\\    
 	10 		&  Acacia shrublands  	& 248		 	&  26     	& 222    	\\    
 	11 		&  Acacia grasslands  	& 305	        &  34     	& 271     	\\    
 	12 		&  Short mopane  		& 181	        &  13     	& 168    	\\    
 	13 		&  Mixed mopane  		& 268 	        &  27      	& 241     	\\    
 	14 		&  Exposed soils  		& 95	        &  6     	& 89     	\\    \hline  
  \end{tabular}
   \caption{\small{Classes and number of test and training samples for Botswana (corresponding to  test/train ratio - 0.90)}}\label{table_Botswana}
\end{table}

\begin{figure}[H]
\captionsetup[subfigure]{labelformat=empty}
\begin{subfigure}{.5\textwidth}
\centering
\includegraphics[scale=.5]{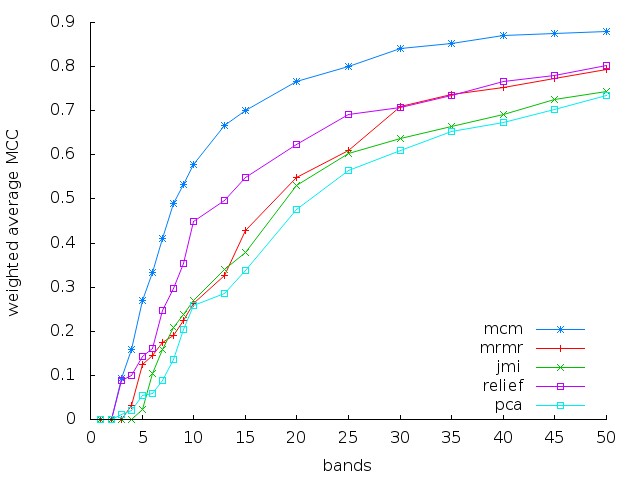}
\caption{\small Weighted average MCC vs number of bands plot for Botswana\\ dataset (test/train ratio - 0.90)}
\end{subfigure}
\begin{subfigure}{.5\textwidth}
\centering
\includegraphics[scale=.5]{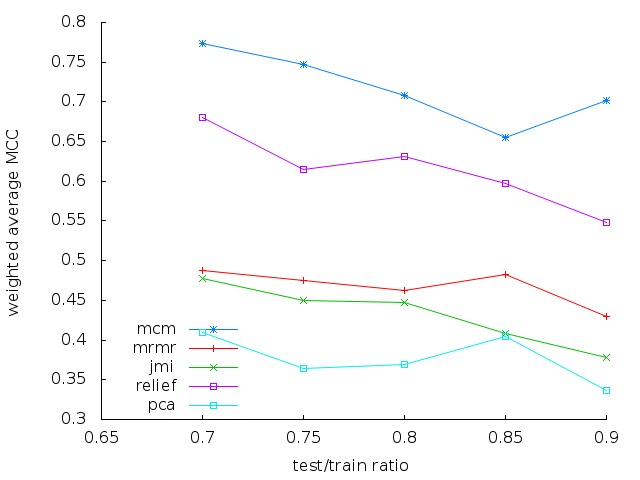}
\caption{\small Weighted average MCC vs test/train ratio plot for Botswana\\ dataset (bands - 15)}
\end{subfigure}
\caption{Botswana dataset}
\label{plot_Botswana}
\end{figure}

Number of bands vs classification accuracy plot is given in figure \ref{plot_Botswana}. Plot of test-train ratio vs classification accuracy generated using first 15 bands is also shown in this figure.

Table \ref{table2_Botswana} shows the top 15 selected bands by different features selection methods.

\begin{table}[H]
  \scriptsize
  \centering
     \noindent \begin{tabular}{ p{.18\columnwidth}  p{.74\columnwidth} }
    \hline
    Band selection method  &  Fifteen selected bands \\  \hline
    MCM    &    23 24 15 34 31 93 85 95 97 70 84 73 87 19 16 \\
    MRMR   &    114 145 71 81 3 6 8 13 14 15 16 17 18 20 21   \\
    JMI    &	114 145 112 4 121 82 71 5 130 131 128 115 107 144 74   \\
    RELIEF &	1 145 143 50 3 5 79 144 142 4 78 62 115 70 71  \\
    PCA    &    140 139 98 141 96 136 99 104 142 93 97 100 106 102 138 \\ \hline
    \end{tabular}
    \caption{\small{comparison between selected bands using different band selection methods
    (test/train ratio-0.90)}}\label{table2_Botswana}
\end{table}

Table \ref{table3_Botswana} depicts the class-wise and weighted average accuracy measure.

\begin{table}[H]
  \scriptsize
  \centering
   \noindent \hspace{-10 pt} \begin{tabular}{ p{.25\columnwidth}  p{.1\columnwidth}  p{.1\columnwidth}  p{.1\columnwidth} p{.1\columnwidth} p{.1\columnwidth} }
    \hline 
    Type 	       			&  MCM      & MRMR 		& JMI        & RELIEF      & PCA  \\    \hline
    Water					& 0.747259  & 0.654967  & 0.6278     & 0.845586    & 0.402704    \\ 
    Hippo grass     		& 0.475178	& 0.294364  & 0.505491   & 0.398122    & 0.204078    \\
    Floodplain grasses1     & 0.855913	& 0.693789  & 0.71528    & 0.6477      & 0.640069    \\
    Floodplain grasses2		& 0.744596	& 0.484395  & 0.528072   & 0.732548    & 0.67        \\
    Reeds1					& 0.837303	& 0.696361  & 0.706814   & 0.752591    & 0.537706    \\
    Riparian				& 0.42237	& 0.0935023 & 0.18776    & 0.480486    & 0.153307    \\
    Firescar2			 	& 0.66327	& -0.00773  & 0.0        & 0.0         & 0.0         \\
    Island interior			& 0.820669	& 0.603341  & 0.580508   & 0.679522    & 0.567368    \\
    Acacia woodlands		& 0.652613	& 0.417259  & 0.163505   & 0.426104    & 0.0         \\
    Acacia shrublands		& 0.82297	& 0.531885  & 0.431715   & 0.854349    & 0.72327     \\
    Acacia grasslands		& 0.626006	& 0.0901463 & 0.125235   & 0.575197    & 0.090038    \\
    Short mopane			& 0.768835	& 0.430352  & 0.384511   & 0.558831    & 0.770241    \\
    Mixed mopane			& 0.631881	& 0.55451   & 0.127159   & 0.39433     & 0.0425411   \\
    Exposed soils			& 0.690334	& 0.618162	& 0.598229   & 0.0		   & 0.0         \\ \hline
	Weighted Average & 0.701248 & 0.429221          & 0.378202   &  0.548067   & 0.336852    \\ \hline     
  \end{tabular}
   \caption{\small{ MCC for different band selection methods for Botswana data-set (corresponding to  test/train ratio - 0.90, bands 15)}}\label{table3_Botswana}
\end{table}

\section{Conclusion}
This paper applies a recently proposed filter feature selection method based on minimizing a tight bound on the VC dimension to the task of hyperspectral image classification. We demonstrate that this feature selection method significantly outperforms state-of-the-art methods in terms classification accuracy that is suitably measured in the presence of a large number of classes. The superior results obtained over different datasets and across a variety of metrics suggests that the proposed method should be the method of choice for this problem. It has not escaped our attention that this method can also be applied to a variety of other high-dimensional classification tasks; we are working on developing modifications of this method for the same.



\begin{thebibliography}{1}

\bibitem{PAL}
Pal, Mahesh, and Giles M. Foody. "Feature selection for classification of hyperspectral data by SVM." Geoscience and Remote Sensing, IEEE Transactions on 48.5 (2010): 2297-2307.

\bibitem{Agustin}
Ifarraguerri, Agustin, and Chein-I. Chang. "Unsupervised hyperspectral image analysis with projection pursuit." Geoscience and Remote Sensing, IEEE Transactions on 38, no. 6 (2000): 2529-2538.

\bibitem{Abhishek_Agarwal}
Agarwal, Abhishek, Tarek El-Ghazawi, Hesham El-Askary, and Jacquline Le-Moigne. "Efficient hierarchical-PCA dimension reduction for hyperspectral imagery." In Signal Processing and Information Technology, 2007 IEEE International Symposium on, pp. 353-356. IEEE, 2007.


\bibitem{SEN_JIA}
Jia, Sen, Zhen Ji, Yuntao Qian, and Linlin Shen. "Unsupervised band selection for hyperspectral imagery classification without manual band removal." Selected Topics in Applied Earth Observations and Remote Sensing, IEEE Journal of 5, no. 2 (2012): 531-543.

\bibitem{Chein_Chang_ID}
Chang, Chein-I., and Su Wang. "Constrained band selection for hyperspectral imagery." Geoscience and Remote Sensing, IEEE Transactions on 44, no. 6 (2006): 1575-1585.

\bibitem{Chein_Chang}
Chang, Chein-I., Qian Du, Tzu-Lung Sun, and Mark LG Althouse. "A joint band prioritization and band-decorrelation approach to band selection for hyperspectral image classification." Geoscience and Remote Sensing, IEEE Transactions on 37, no. 6 (1999): 2631-2641.

\bibitem{Baofeng_Guo}
Guo, Baofeng, Steve R. Gunn, R. I. Damper, and J. D. B. Nelson. "Band selection for hyperspectral image classification using mutual information." Geoscience and Remote Sensing Letters, IEEE 3, no. 4 (2006): 522-526. 

\bibitem{FS_Jayadeva}
Jayadeva, Batra, Sanjit S., and Siddharth Sabharwal. "Feature Selection through Minimization of the VC dimension." arXiv preprint arXiv:1410.7372 (2014).

\bibitem{Farid_Melgani}
Melgani, Farid, and Lorenzo Bruzzone. "Classification of hyperspectral remote sensing images with support vector machines." Geoscience and Remote Sensing, IEEE Transactions on 42, no. 8 (2004): 1778-1790.

\bibitem{Adolfo_Martinez}
Martínez-Usó, Adolfo, Filiberto Pla, José Martínez Sotoca, and Pedro García-Sevilla. "Clustering-based hyperspectral band selection using information measures." Geoscience and Remote Sensing, IEEE Transactions on 45, no. 12 (2007): 4158-4171. 

\bibitem{Guyon}
Guyon, Isabelle, Jason Weston, Stephen Barnhill, and Vladimir Vapnik. "Gene selection for cancer classification using support vector machines." Machine learning 46, no. 1-3 (2002): 389-422.
 
 \bibitem{Hall}
Hall, Mark A., and Lloyd A. Smith. "Feature subset selection: a correlation based filter approach." (1997).

\bibitem{Peng}
Peng, Hanchuan, Fulmi Long, and Chris Ding. "Feature selection based on mutual information criteria of max-dependency, max-relevance, and min-redundancy." Pattern Analysis and Machine Intelligence, IEEE Transactions on 27, no. 8 (2005): 1226-1238.

\bibitem{Breiman}
Breiman, Leo. "Random forests." Machine learning 45, no. 1 (2001): 5-32.

\bibitem{Gavin_Brown}
Brown, Gavin, Adam Pocock, Ming-Jie Zhao, and Mikel Luján. "Conditional likelihood maximisation: a unifying framework for information theoretic feature selection." The Journal of Machine Learning Research 13, no. 1 (2012): 27-66.

\bibitem{Yang}
Yang, Howard Hua, and John E. Moody. "Data Visualization and Feature Selection: New Algorithms for Nongaussian Data." In NIPS, pp. 687-702. 1999.

\bibitem{Fleuret}
Fleuret, François. "Fast binary feature selection with conditional mutual information." The Journal of Machine Learning Research 5 (2004): 1531-1555.

\bibitem{Kira}
Kira, Kenji, and Larry A. Rendell. "The feature selection problem: Traditional methods and a new algorithm." In AAAI, vol. 2, pp. 129-134. 1992.

\bibitem{Classifier_MCM}
Jayadeva. "Learning a hyperplane classifier by minimizing an exact bound on the VC dimension." NEUROCOMPUTING 149 (2015): 683-689.

\bibitem{Classifier_MCM_arxiv}
Jayadeva. "Learning a hyperplane classifier by minimizing an exact bound on the VC dimension." arXiv preprint arXiv:1408.2803 (2014).

\bibitem{Regressor_MCM}
Jayadeva, Chandra, Suresh, Sanjit S. Batra, and Siddarth Sabharwal. "Learning a hyperplane regressor through a tight 
bound on the VC dimension." Neurocomputing (2015).

\bibitem{Regressor_MCM_arxiv}
Jayadeva, Chandra, Suresh, Siddarth Sabharwal, and Sanjit S. Batra. "Learning a hyperplane regressor by minimizing an exact bound on the VC dimension." arXiv preprint arXiv:1410.4573 (2014).

\bibitem{LibSVM}
Chang, Chih-Chung, and Chih-Jen Lin. "LIBSVM: A library for support vector machines." ACM Transactions on Intelligent Systems and Technology (TIST) 2.3 (2011): 27.
 
\end{thebibliography}
%

\end{document}